\documentclass{article}

\usepackage{microtype}
\usepackage{graphicx}
\usepackage{subcaption}
\usepackage{booktabs} 

\usepackage{hyperref}

\usepackage[preprint]{icml2026}

\usepackage{amsmath}
\usepackage{amssymb}
\usepackage{mathtools}
\usepackage{amsthm}

\usepackage[capitalize,noabbrev]{cleveref}

\theoremstyle{plain}

\theoremstyle{definition}

\theoremstyle{remark}

\usepackage[textsize=tiny]{todonotes}

\icmltitlerunning{Adversarial Information Separation}

\usepackage{multirow}
\usepackage{enumitem}

\begin{document}

\twocolumn[
  \icmltitle{Refining the Information Bottleneck via Adversarial Information Separation} 



  \icmlsetsymbol{equal}{*}

  \begin{icmlauthorlist}
    \icmlauthor{Shuai Ning}{equal,gibm}
    \icmlauthor{Zhenpeng Wang}{equal,uic}
    \icmlauthor{Lin Wang}{uic,qcl}
    \icmlauthor{Bing Chen}{waterloo}
    \icmlauthor{Shuangrong Liu}{uic}
    \icmlauthor{Xu Wu}{gibm}
    \icmlauthor{Jin Zhou}{swu}
    \icmlauthor{Bo Yang}{qcl}
  \end{icmlauthorlist}

  \icmlaffiliation{uic}{Shandong Key Laboratory of Ubiquitous Intelligent Computing, University of Jinan, Shandong, China}
  \icmlaffiliation{gibm}{Shandong Provincial Key Laboratory of Green and Intelligent Building Materials, University of Jinan, Shandong, China}
  \icmlaffiliation{qcl}{Quan Cheng Laboratory, Jinan, China}
  \icmlaffiliation{waterloo}{David R. Cheriton School of Computer Science, University of Waterloo, Waterloo, Canada}
  \icmlaffiliation{swu}{School of Artificial Intelligence, Shandong Women’s University, Jinan, China}

  \icmlcorrespondingauthor{Lin Wang}{wangplanet@gmail.com}

  \icmlkeywords{Information Bottleneck, Adversarial Learning, Representation Learning, Out-of-Distribution Generalization, AI for Science}

  \vskip 0.3in
]



\printAffiliationsAndNotice{\icmlEqualContribution}

\begin{abstract}
Generalizing from limited data is particularly critical for models in domains such as material science, where task-relevant features in experimental datasets are often heavily confounded by measurement noise and experimental artifacts.
Standard regularization techniques fail to precisely separate meaningful features from noise, while existing adversarial adaptation methods are limited by their reliance on explicit separation labels.
To address this challenge, we propose the Adversarial Information Separation Framework (AdverISF), which isolates task-relevant features from noise without requiring explicit supervision.
AdverISF introduces a self-supervised adversarial mechanism to enforce statistical independence between task-relevant features and noise representations.
It further employs a multi-layer separation architecture that progressively recycles noise information across feature hierarchies to recover features inadvertently discarded as noise, thereby enabling finer-grained feature extraction.
Extensive experiments demonstrate that AdverISF outperforms state-of-the-art methods in data-scarce scenarios. 
In addition, evaluations on real-world material design tasks show that it achieves superior generalization performance.
\end{abstract}


\section{Introduction}

Generalization performance measures a model's predictive robustness on unseen data, such as test sets or real-world scenarios~\cite{goodfellow2016deep,xu2012robustness}. A well-generalized model effectively disentangles underlying patterns from noise, ensuring consistent predictions across dynamic environments. This adaptability is paramount in scientific and engineering domains ~\cite{wang2023scientific,liu2021towards}, where the objective is to handle future observations rather than merely fit training distributions.

While conventional deep learning thrives on massive datasets, data scarcity in practical applications hinders the development of highly generalizable models. Strategies such as large-scale pre-training~\cite{radford2021learning} and meta-learning~\cite{finn2017model} have emerged to address this issue. These methods leverage large datasets accumulated from related tasks for pre-training or optimizing learning strategies. They have achieved remarkable success in data-rich fields such as computer vision and natural language processing~\cite{devlin2019bert,deng2009imagenet}.

\begin{figure}
    \centering
    \includegraphics[width=0.7\linewidth]{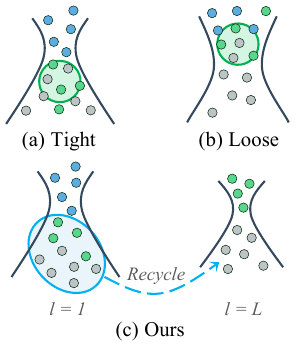}
    \caption{(a) A tight bottleneck retains dominant features (blue) but discards subtle features (green). (b) A loose bottleneck captures subtle features but leaks significant noise (gray). (c) Ours employs a layer-wise mechanism to recycle subtle features that would otherwise be erroneously discarded, effectively preserving meaningful information while filtering out noise.}
    \label{fig:Principles}
\end{figure}

However, applying these large-scale data paradigms to specialized domains such as materials science remains a formidable challenge~\cite{xu2023small,dou2023machine}. The primary obstacle is the substantial heterogeneity of data formats and distributional patterns. Consider cementitious materials as an example. Data originating from different studies exhibit high variance in compositional makeup and physicochemical characteristics due to diverse raw material sources and complex reaction mechanisms~\cite{li2022machine,juenger2019supplementary}. These pronounced disparities make it exceptionally difficult to integrate multi-source data for constructing high-quality benchmarks and pre-training general-purpose models.

In the absence of large-scale data, the Information Bottleneck (IB)~\cite{tishby1999information,pmlr-v202-kawaguchi23a,NEURIPS2021_1c336b80} theory offers a robust framework for extracting essential features from noise. Following the principle of minimal sufficient statistics, IB seeks an optimal representation $Z$ that maximizes information about the target $Y$ while minimizing redundancy from the input $X$. This trade-off mechanism compels the model to eliminate task-irrelevant confounders, thereby retaining only features with genuine generalization capability. The objective is formalized as:
\begin{equation}
    \min \mathcal{L}_{IB} = I(X; Z) - \beta I(Y; Z),
\end{equation}
where $\beta$ is a Lagrange multiplier balancing compression and prediction. While theoretically elegant, the direct application of IB to deep neural networks has been impeded by the intractability of computing mutual information in complex high-dimensional spaces~\cite{yu2021information,alemi2017deep,NEURIPS2024_efa9e3ed,chow2005estimating}.

The Variational Information Bottleneck (VIB)~\cite{alemi2017deep} addresses this intractability by employing variational inference to construct differentiable upper bounds. However, VIB relies on minimizing the KL divergence between learned representations and a static prior for information compression~\cite{choi2024conditional,8680020,pan2021disentangled,pmlr-v235-wang24cm,hu-etal-2024-representation}. This statistical compression strategy often fails to disentangle noise in complex high-dimensional feature spaces. It inadvertently discards secondary yet informative features alongside redundancy, limiting generalization performance, as visualized in Figure 1.

Xie et al.~\cite{xie2024information} addressed this information loss by introducing an Information Retention framework. They argued that features conventionally deemed redundant can be crucial for out-of-distribution (OOD) generalization. However, their approach remains grounded in the VIB paradigm and relies on statistical constraints to recover these secondary features. Consequently, the method is limited in its ability to accurately distinguish informative signals from complex high-dimensional noise.

Adversarial learning offers a compelling alternative to statistical compression. 
By employing discriminators to enforce distribution matching, adversarial methods can effectively disentangle feature components. 
Pioneering works in domain adaptation~\cite{ganin2016domain,Pei_Cao_Long_Wang_2018,NEURIPS2018_ab88b157} and robust learning~\cite{kim2021distilling,NEURIPS2019_e2c420d9,9719702} have demonstrated that task-discriminative features can be separated from nuisances through adversarial min-max games. 
However, these strategies fundamentally rely on explicit supervision, such as domain labels or robust annotations, to define the separation boundary\cite{pmlr-v119-liang20a,10.1007/978-3-030-30645-8_36,pmlr-v97-zhao19a}. 
In specialized domains with limited data, such labels are typically unavailable.
Each experiment produces data from a single setup, lacking the multi-domain labels required for adversarial domain adaptation.
The standard supervision-based adversarial strategy is therefore not directly applicable.

This raises our central question. \textit{How can we leverage adversarial mechanisms to actively separate task-relevant features from noise without explicit supervision, thereby enhancing generalization when training data is scarce?}

\textbf{Contribution} To address this question, we propose an Adversarial Information Separation Framework (AdverISF) that employs dual-channel encoders to map inputs into independent spaces for task-relevant features and noise. We leverage an adversarial mechanism that enforces statistical independence by distinguishing paired from shuffled representations, and extend this to a multi-layer architecture for progressive feature refinement. The main contributions of this work are as follows.

\begin{enumerate}
    \item We propose a self-supervised adversarial separation mechanism that enforces statistical independence between task-relevant features and noise by distinguishing joint distributions from product-of-marginals. Unlike prior adversarial methods requiring domain labels or robust annotations, our approach achieves effective separation in single-domain scenarios without explicit signal-noise supervision.
    
    \item We introduce a multi-layer separation architecture that cascades separation blocks and recycles noise representations to sequentially extract both dominant and subtle features. This overcomes the uniform compression limitation of existing information bottleneck methods, which suppress subtle yet generalizable signals alongside noise.
    
    \item Extensive experiments across synthetic benchmarks and real-world material science applications demonstrate substantial improvements in data-scarce regimes. On a real-world composite cement design task, AdverISF achieves $R^2=0.897$ on unseen formulations, significantly outperforming competing methods.
\end{enumerate}

\begin{figure*}[tb]
    \centering
    \includegraphics[width=0.95\linewidth]{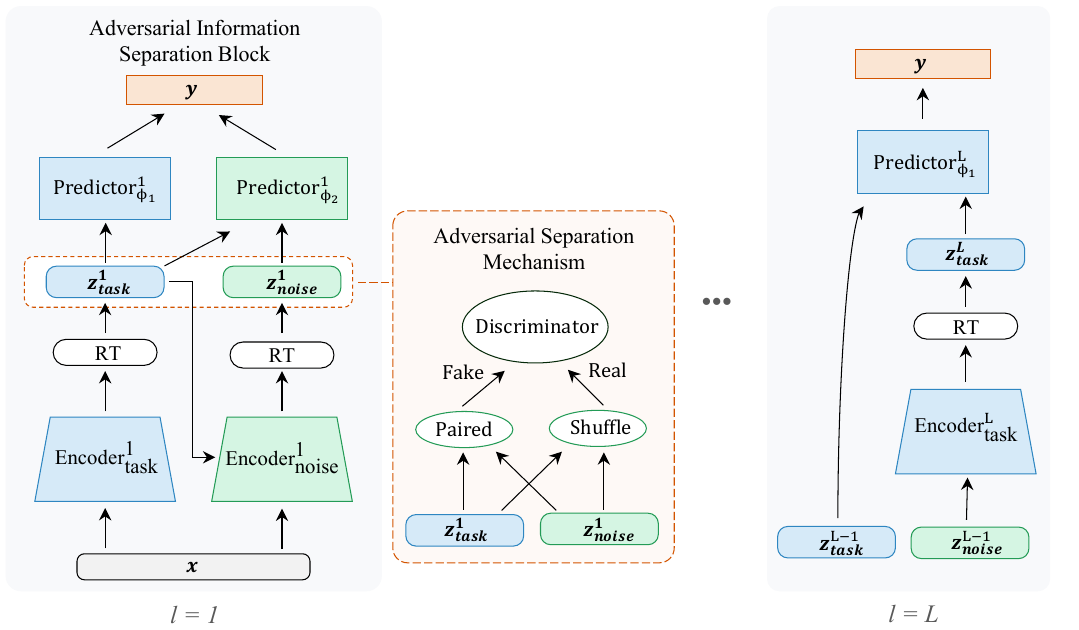}
    \caption{
        \textbf{Schematic of the AdverISF.} 
        The architecture adopts a \textbf{Multi-layer Separation} strategy ($l=1$ to $L$) for progressive feature refinement. 
        At each level, the \textbf{Adversarial Information Separation Block} utilizes dual encoders to decompose the input into task-relevant features ($z_{task}$) and noise representations ($z_{noise}$). Reparameterization (RT) is applied to enable differentiable sampling from the latent distributions.
        A central \textbf{Adversarial Separation Mechanism} enforces statistical independence between these latent codes via a minimax game, where a discriminator $D$ distinguishes between paired (joint) and shuffled (marginal) distributions. 
        The noise output from the previous layer serves as input to the subsequent layer to capture subtle features $z_{subtle}$ (represented by $z_{task}^L$ in Layer $L$).
    }
    \label{fig:AdverISF}
\end{figure*}

\section{Related Works}

\textbf{Learning from Multiple Tasks and Domains}
Generalization strategies often leverage external knowledge. Representative approaches include large-scale pre-training~\cite{deng2009imagenet, radford2021learning}, meta-learning~\cite{finn2017model}, and domain-specific models~\cite{lin2023evolutionary}. However, these methods rely on massive datasets or diverse source domains. They are often inapplicable to specialized scientific fields where data is inherently scarce and high-quality auxiliary datasets are unavailable.

\textbf{Information Bottleneck Methods}
The Information Bottleneck (IB) principle balances prediction accuracy with compression~\cite{tishby1999information}. The Variational IB (VIB) applied this to deep networks~\cite{alemi2017deep}, with subsequent extensions for robustness and domain adaptation~\cite{kuang2023improving,li2022invariant}. Despite these advancements, most IB methods rely on passive statistical regularization against a fixed prior. They often fail to actively distinguish subtle task-relevant features from structured noise in complex single-domain scenarios.

\textbf{Adversarial Representation and Disentanglement}
Disentangled representation learning\cite{NEURIPS2018_1ee3dfcd,10.5555/3692070.3692839,NEURIPS2023_8e63972d,pmlr-v139-zhu21f}, such as $\beta$-VAE~\cite{higgins2017beta} and FactorVAE~\cite{kim2018disentangling}, enforces factorization to enhance interpretability. However, these objectives are typically unsupervised and task-agnostic. They treat all factors equally and lack a mechanism to differentiate task-relevant signals from nuisances. This limits their effectiveness in supervised settings where specific features must be prioritized.

Adversarial learning offers a powerful mechanism for structural constraints. Methods like ANICA~\cite{brakel2017learning} employ discriminators to enforce statistical independence between features. While effective for disentanglement, they operate in an unsupervised manner and cannot identify task relevance. Conversely, domain adaptation methods~\cite{Pei_Cao_Long_Wang_2018,NEURIPS2018_ab88b157} like DANN~\cite{ganin2016domain} use adversarial training to explicitly separate discriminative features from variations. However, these approaches depend on explicit supervision such as domain labels.

\textbf{Challenge}
A critical gap remains in specialized, data-scarce domains.
External knowledge approaches require auxiliary data that is often unavailable, while domain adaptation methods depend on explicit domain labels.
Moreover, existing methods based on statistical independence, such as ANICA, treat all feature components equally and fail to prioritize task-relevant signals over noise.
The fundamental challenge is achieving explicit feature separation without domain labels or auxiliary supervision.

\section{Methodology}
The Adversarial Information Separation Framework (AdverISF) treats information separation as an active adversarial game between two competing objectives. Rather than relying on passive compression via statistical regularization, we explicitly model the competition between task-relevant feature preservation and noise isolation.
We formalize this framework in the following stages. Section 3.1 establishes the noisy observation setting, where observed inputs $x$ confound task-relevant features $z_{task}$ with noise $z_{noise}$. Sections 3.2 and 3.3 present a dual-branch architecture that explicitly separates these two representations. The key innovation lies in an adversarial mechanism that minimizes mutual information $I(z_{task}; z_{noise})$ through a minimax game. Section 3.4 then demonstrates how this approach scales to multi-layer hierarchies for progressive feature refinement.

\subsection{Problem Formulation}
\label{sec:formulation}
In standard supervised learning, a model learns a mapping from input space $\mathcal{X}$ to target space $\mathcal{Y}$: 
\begin{equation} 
    y = f(x) + \epsilon,
\end{equation}
where $x \in \mathcal{X}$ is the observed input (or feature vector), $y \in \mathcal{Y}$ is the corresponding target (or label), and $\epsilon$ denotes irreducible error. This formulation assumes that the $x$ directly encodes all task-relevant information. In many real-world scenarios, however, the target $y$ depends on latent features $z_{task}$ that are not directly observable. Instead, we observe $x$ through a potentially complex observation process. This leads to a decomposed generative model:
\begin{equation}
    y = f(z_{task}) + \epsilon, \quad x = \mathcal{O}(z_{task}),
\end{equation}
where $\mathcal{O}(\cdot)$ denotes the observation function. When $\mathcal{O}$ is an identity mapping ($x = z_{task}$), the model has direct access to task-relevant features. In many practical domains, however, the observation process is corrupted by environmental interference, instrumental noise, and other confounding factors. The observed input $x$ then becomes a mixture of task-relevant and task-irrelevant information:
\begin{equation}
    x = \mathcal{G}(z_{task}, z_{noise}),
\end{equation}
where $\mathcal{G}(\cdot)$ represents the noisy observation mechanism and $z_{noise}$ denotes nuisance factors that are irrelevant to the prediction target $y$. This confounding poses a key challenge for learning. Standard models trained on $x$ may inadvertently rely on spurious correlations with $z_{noise}$, leading to poor generalization when the noise distribution shifts.

Given this noisy observation setting, the IB framework aims to learn a compressed representation $z$ that maximizes information about the target $y$ while minimizing information from the input $x$. This approach implicitly treats all task-relevant features as equally important and uniformly compressible. In practice, IB methods such as VIB implement this compression through a global statistical penalty that applies uniform pressure across all representation dimensions. Specifically, VIB regularizes the learned representation by minimizing its KL divergence to a fixed prior.

This homogeneity assumption rarely holds in complex domains. Task-relevant features typically exhibit hierarchical importance, where some components contribute more critically to the prediction than others. We can decompose $z_{task}$ into two types of components. The first type, denoted $z_{dominant}$, consists of salient, primary features that are highly predictive of $y$. The second type, denoted $z_{subtle}$, represents fine-grained, secondary information that are crucial for generalization but exhibit subtle variations and are easily overshadowed by noise.

However, this uniform compression strategy causes VIB to struggle in distinguishing subtle meaningful features $z_{subtle}$ from complex noise patterns $z_{noise}$. Both manifest as weak variations that are easily overwhelmed by the regularization strength of the global KL penalty. As a result, $z_{subtle}$ is frequently suppressed alongside noise during compression. This leads to information loss that is critical for out-of-distribution generalization.

\subsection{Overview of AdverISF}
To address the limitations of uniform statistical compression in VIB, we propose the Adversarial Information Separation Framework (AdverISF). Rather than relying on passive compression through a single encoder with a global KL penalty, AdverISF actively separates task-relevant features from noise using dual encoders. These encoders explicitly allocate $z_{task}$ and $z_{noise}$ into distinct latent spaces. An adversarial mechanism enforces this separation by minimizing the mutual information between the two representations. 

The framework is built upon a fundamental building block called the Adversarial Information Separation Block, which we describe in Section~\ref{sec:block}. This block can be cascaded into a multi-layer architecture for progressive refinement of hierarchical features, as detailed in Section~\ref{sec:hierarchy}. Figure~\ref{fig:AdverISF} illustrates the overall architecture.

\subsection{Adversarial Information Separation Block}
\label{sec:block}

The Adversarial Information Separation Block serves as the fundamental building block of the model. It decomposes input data into dual latent spaces through three key sub-modules. The Feature Extractor captures task-relevant information, while the Noise Extractor models residual variations and confounding factors. A core challenge in this decomposition is to prevent the two extractors from encoding overlapping information. We address this through an Adversarial Separation Mechanism that enforces statistical independence between the learned representations. This mechanism acts as a regularizer to ensure clean separation and avoid information leakage across the two spaces.

\subsubsection{Feature Extractor}
The Feature Extractor follows the VIB framework. It distills a compact representation $z_{task}$ that retains maximal information about the target $y$ while compressing irrelevant details from the input $x$. This branch consists of two components. A Feature Encoder maps inputs to the latent space, and a Task Predictor estimates the target variable.

Our objective is to maximize $I(z_{task}; y) - \beta_{task} \cdot I(z_{task}; x)$. Since direct optimization of mutual information terms is computationally intractable, we employ variational approximation. The Feature Encoder models the posterior distribution $p_{\theta}(z_{task}|x)$, and the Task Predictor approximates the likelihood $q_{\psi_1}(y|z_{task})$. This yields the following loss:
\begin{equation}
\begin{aligned}
    \mathcal{L}_{task} = & \, \mathbb{E}_{x}[\mathbb{E}_{z_{task} \sim p_{\theta}(z_{task}|x)} [-\log q_{\psi_1}(y|z_{task})]] \\
    & + \beta_{task} \cdot D_{KL}[p_{\theta}(z_{task}|x) \parallel r(z_{task})],
\end{aligned}
\end{equation}
where $\beta_{task}$ controls the trade-off between prediction accuracy and information compression.

\subsubsection{Noise Extractor Branch}

The Noise Extractor follows a similar architecture but operates conditionally to capture residual information. It consists of a Noise Encoder and a Joint Predictor. The Noise Encoder extracts latent factors $z_{noise}$ from the input $x$ that are complementary to the already extracted $z_{task}$. It models the conditional distribution $p_{\phi}(z_{noise}|x, z_{task})$, taking both the original input and the task features as inputs. The Joint Predictor $q_{\psi_2}(y|z_{task}, z_{noise})$ then utilizes both latent codes to predict the target. This design ensures that the combination of $z_{task}$ and $z_{noise}$ captures the full information content of $x$ required for fitting the training data.

The objective of this branch is to maximize the conditional mutual information $I(z_{noise}; y \mid z_{task}) - \beta_{noise} \cdot I(z_{noise}; x \mid z_{task})$. The corresponding loss function is:
\begin{equation}
\begin{split}
    & \mathcal{L}_{noise} = \mathbb{E}_{x, z_{task}} \Big[ \mathbb{E}_{z_{noise} \sim p_{\phi}(z_{noise}|x, z_{task})} \\
    &\qquad \qquad [-\log q_{\psi_2}(y|z_{task}, z_{noise})] \Big] \\  & \quad + \beta_{noise} \cdot D_{KL}[p_{\phi}(z_{noise}|x, z_{task}) \parallel r(z_{noise})],
\end{split}
\end{equation}
where $\beta_{noise}$ controls the compression pressure on $z_{noise}$ and is set to a small value. This minimal compression allows the branch to retain fine-grained details and noise patterns that might be discarded by the Feature Extractor.

\textbf{Role of Joint Prediction}
Why maximize the mutual information between the noise representation $z_{noise}$ and the target $y$ when noise is typically task-irrelevant? This design choice is critical for the effectiveness of the adversarial game. Without supervision, the Noise Encoder could degenerate into generating generic Gaussian noise unrelated to the input $x$. In this trivial case, $z_{task}$ and $z_{noise}$ would be naturally independent, satisfying the adversarial constraint without enforcing any meaningful purification of $z_{task}$. To prevent this degeneration, we employ the Joint Predictor to compel $z_{noise}$ to capture residual information that is predictive of $y$. This includes both subtle task-relevant features and spurious correlations. The result is a strong adversary that forces the subsequent adversarial separation mechanism to perform non-trivial disentanglement, thereby ensuring the purity of $z_{task}$.

\subsubsection{Adversarial Separation Mechanism}

The Adversarial Separation Mechanism enforces strict separation between the information encoded in $z_{task}$ and $z_{noise}$. The mathematical objective is to minimize the mutual information between the two latent representations, that is, to minimize $I(z_{noise}; z_{task})$.
This design is inspired by the work of Ganin et al. on Domain-Adversarial Neural Networks (DANN)~\cite{ganin2016domain}. DANN uses adversarial training to align feature distributions across different domains. A discriminator is trained to classify explicit domain labels such as source versus target, while the feature extractor attempts to confuse this classifier. Our scenario differs fundamentally. We lack explicit labels distinguishing signal from noise within a single input $x$. The standard label-based adversarial strategy is therefore inapplicable. 

We propose a distribution-based adversarial strategy. Instead of classifying labels, our discriminator distinguishes between the joint distribution and the product of marginal distributions. For a given batch of inputs $X=\{x_1, \dots, x_n\}$, the encoders generate paired representations $Z_{paired} = \{(z_{j, task}, z_{j, noise})\}_{j=1}^n$. These serve as samples from the joint distribution $p(z_{task}, z_{noise})$. We then apply a random permutation to the noise representations within the batch to construct a shuffled set $Z_{shuffled} = \{(z_{j, task}, z_{\pi(j), noise})\}_{j=1}^n$, where $\pi$ denotes a random index permutation. This process breaks the instance-wise correspondence and effectively approximates samples from the product of marginals $p(z_{task})p(z_{noise})$. We provide a theoretical justification for this independence enforcement mechanism in Appendix~\ref{app:theory}.

We use WGAN-GP~\cite{arjovsky2017wasserstein,gulrajani2017improved} for adversarial training. The discriminator $D$ estimates the Wasserstein distance between paired and shuffled samples, which the encoders aim to minimize. The objective with gradient penalty is defined as:

\begin{equation}
\begin{aligned}
\mathcal{L}_{D} = & \;\mathbb{E}_{shuffled} [D(z_{noise}, z'_{task})] \\ & \qquad - \mathbb{E}_{paired} [D(z_{noise}, z_{task})]
 \\ & \qquad  + \gamma \mathbb{E}_{\hat{z}} [(\| \nabla_{\hat{z}} D(\hat{z}) \|_2 - 1)^2],
\end{aligned}
\end{equation}

where $\hat{z}$ denotes the random interpolation between paired and shuffled samples, and $\gamma$ is the penalty coefficient.

\subsubsection{Total Optimization Objective}

Integrating the objectives from the feature extractor, noise extractor, and adversarial mechanism, the final training objective is formulated as a minimax game.
\begin{equation}
    \min_{\theta, \phi, \psi} \max_{D} \mathcal{L}_{total} =   \mathcal{L}_{task} + \mathcal{L}_{noise} + \lambda_{adv} \cdot \mathcal{L}_{adv},
\end{equation}
Here $\lambda_{adv}$ are hyperparameter balancing the contributions of task prediction, noise reconstruction, and adversarial separation. The parameters $\theta$ and $\phi$ correspond to the Feature Encoder and Noise Encoder respectively, and $\psi = \{\psi_1, \psi_2\}$ parameterizes the Task Predictor and Joint Predictor.

\begin{table}[tb]
\centering
\renewcommand{\arraystretch}{1.05}
\setlength{\tabcolsep}{1.5pt}
\caption{Generalization performance ($R^2$ $\uparrow$) on the \textbf{AEP Dataset}. Mean $\pm$ std over 10 seeds. Best results in \textbf{bold}.}
\label{tab:r2_aep}
\resizebox{\columnwidth}{!}{
\begin{tabular}{lcccc}
\toprule
\textbf{Model} & \textbf{10\%} & \textbf{20\%} & \textbf{50\%} & \textbf{70\%} \\
\midrule
MLP & 0.322 {\scriptsize $\pm$ 0.08} & 0.444 {\scriptsize $\pm$ 0.02} & 0.564 {\scriptsize $\pm$ 0.02} & 0.558 {\scriptsize $\pm$ 0.02} \\
VIB & 0.426 {\scriptsize $\pm$ 0.05} & 0.490 {\scriptsize $\pm$ 0.05} & 0.578 {\scriptsize $\pm$ 0.03} & \textbf{0.604 {\scriptsize $\pm$ 0.03}} \\
infoR--LSF & 0.346 {\scriptsize $\pm$ 0.06} & 0.483 {\scriptsize $\pm$ 0.05} & 0.537 {\scriptsize $\pm$ 0.03} & 0.571 {\scriptsize $\pm$ 0.03} \\
\midrule
Ours (Joint) & 0.430 {\scriptsize $\pm$ 0.07} & \textbf{0.501 {\scriptsize $\pm$ 0.04}} & \textbf{0.599 {\scriptsize $\pm$ 0.03}} & 0.597 {\scriptsize $\pm$ 0.03} \\
Ours (Two-Stage) & \textbf{0.431 {\scriptsize $\pm$ 0.08}} & 0.416 {\scriptsize $\pm$ 0.01} & 0.423 {\scriptsize $\pm$ 0.08} & 0.461 {\scriptsize $\pm$ 0.12} \\
\bottomrule
\end{tabular}
}
\end{table}

\begin{table}[tb]
\centering
\renewcommand{\arraystretch}{1.05}
\setlength{\tabcolsep}{1.5pt}
\caption{Generalization performance ($R^2$ $\uparrow$) on the \textbf{Concrete Dataset}. Mean $\pm$ std over 10 seeds. Best results in \textbf{bold}.}
\label{tab:r2_concrete}
\resizebox{\columnwidth}{!}{
\begin{tabular}{lccccc}
\toprule
\textbf{Model} & \textbf{30} & \textbf{50} & \textbf{70} & \textbf{100} & \textbf{500} \\
\midrule
MLP & 0.207 {\scriptsize $\pm$ 0.09} & 0.381 {\scriptsize $\pm$ 0.04} & 0.581 {\scriptsize $\pm$ 0.02} & 0.649 {\scriptsize $\pm$ 0.01} & \textbf{0.886 {\scriptsize $\pm$ 0.00}} \\
VIB & 0.270 {\scriptsize $\pm$ 0.07} & 0.386 {\scriptsize $\pm$ 0.05} & 0.589 {\scriptsize $\pm$ 0.03} & 0.641 {\scriptsize $\pm$ 0.03} & 0.877 {\scriptsize $\pm$ 0.01} \\
infoR--LSF & 0.269 {\scriptsize $\pm$ 0.05} & 0.451 {\scriptsize $\pm$ 0.03} & 0.585 {\scriptsize $\pm$ 0.02} & 0.658 {\scriptsize $\pm$ 0.02} & 0.882 {\scriptsize $\pm$ 0.00} \\
\midrule
Ours (Joint) & 0.496 {\scriptsize $\pm$ 0.02} & 0.638 {\scriptsize $\pm$ 0.01} & 0.662 {\scriptsize $\pm$ 0.01} & \textbf{0.674 {\scriptsize $\pm$ 0.01}} & 0.815 {\scriptsize $\pm$ 0.02} \\
Ours (Two-Stage) & \textbf{0.526 {\scriptsize $\pm$ 0.02}} & \textbf{0.643 {\scriptsize $\pm$ 0.01}} & \textbf{0.685 {\scriptsize $\pm$ 0.01}} & 0.669 {\scriptsize $\pm$ 0.01} & 0.829 {\scriptsize $\pm$ 0.00} \\
\bottomrule
\end{tabular}
}
\end{table}

\subsection{Multi-layer Separation Architecture}
\label{sec:hierarchy}

As discussed in Sec.~\ref{sec:formulation}, task-relevant features in complex domains exhibit a hierarchical structure, ranging from primary feature $z_{dominant}$ to subtle factors $z_{subtle}$. A single Adversarial Information Separation Block may struggle to capture this complete hierarchy.
Driven by the strong adversarial constraint, the model adopts a conservative strategy that prioritizes the most discriminative factors to ensure feature purity. While effective for separation, this purity-completeness trade-off may leave less prominent but critical subtle features entangled within the noise representation.

We address this limitation by extending the single-block design into a Multi-layer Separation Architecture. This framework adopts a progressive purification strategy by cascading multiple Adversarial Information Separation Blocks. Crucially, the noise representation output from the $l$-th layer is recycled as input to the $(l+1)$-th layer, enabling the recovery of subtle features that may have been misclassified as noise. We provide an information-theoretic justification for the hierarchical recovery mechanism in Appendix~\ref{app:multilayer_theory}.

In the first layer, the block takes the raw input $x$ and extracts the most salient features as $z_{task}^{1}$. All remaining information, including $z_{subtle}$ and actual noise, is pushed into $z_{noise}^{1}$. In the subsequent layer, the second block operates on $z_{noise}^{1}$ to extract the secondary features $z_{task}^{2}$ and isolate the remaining residuals in $z_{noise}^{2}$. This process repeats recursively. Finally, the comprehensive task representation is constructed by aggregating the feature codes from all layers. This ensures that both dominant and subtle information are preserved for the final prediction.

\section{Experiments}
We evaluate AdverISF on synthetic and real-world datasets to investigate three core questions. 
(1) Can AdverISF achieve superior generalization performance in data-scarce and out-of-distribution scenarios compared to state-of-the-art baselines? (Sec.~\ref{sec:comparison} and Sec.~\ref{sec:realworld})
(2) Do the proposed adversarial separation mechanism and the multi-layer architecture effectively contribute to the model's performance? (Sec.~\ref{sec:ablation})
(3) How does the model behavior vary with hyperparameter settings, and does it successfully learn disentangled representations as theoretically intended? (Sec.~\ref{sec:hyperparam})

\begin{table}[tb]
\centering
\setlength{\tabcolsep}{1.5pt}
\caption{Generalization performance ($R^2$ $\uparrow$) on the \textbf{Synthetic Dataset}. Mean $\pm$ std over 10 seeds. Best results in \textbf{bold}.}
\label{tab:r2_synthetic}
\resizebox{\columnwidth}{!}{
\begin{tabular}{lcccc}
\toprule
\textbf{Model} & \textbf{10\%} & \textbf{20\%} & \textbf{50\%} & \textbf{70\%} \\
\midrule
MLP & 0.201 {\scriptsize $\pm$ 0.04} & 0.230 {\scriptsize $\pm$ 0.01} & 0.421 {\scriptsize $\pm$ 0.01} & 0.468 {\scriptsize $\pm$ 0.03} \\
VIB & 0.159 {\scriptsize $\pm$ 0.07} & 0.147 {\scriptsize $\pm$ 0.04} & 0.393 {\scriptsize $\pm$ 0.03} & 0.478 {\scriptsize $\pm$ 0.06} \\
infoR--LSF & 0.165 {\scriptsize $\pm$ 0.05} & 0.147 {\scriptsize $\pm$ 0.04} & 0.406 {\scriptsize $\pm$ 0.03} & 0.435 {\scriptsize $\pm$ 0.03} \\
\midrule
Ours (Joint) & 0.179 {\scriptsize $\pm$ 0.06} & \textbf{0.254 {\scriptsize $\pm$ 0.06}} & \textbf{0.423 {\scriptsize $\pm$ 0.05}} & 0.355 {\scriptsize $\pm$ 0.08} \\
Ours (Two-Stage) & \textbf{0.251 {\scriptsize $\pm$ 0.07}} & 0.253 {\scriptsize $\pm$ 0.09} & 0.420 {\scriptsize $\pm$ 0.04} & \textbf{0.565 {\scriptsize $\pm$ 0.06}} \\
\bottomrule
\end{tabular}
}
\vspace{-0.3cm}
\end{table}

\begin{table}[tb]
\centering
\setlength{\tabcolsep}{9pt}
\caption{Classification accuracy on \textbf{CIFAR-10}. Mean accuracy over 10 seeds. Best results in \textbf{bold}.}
\label{tab:cifar10_acc}
\resizebox{\columnwidth}{!}{
\begin{tabular}{lccccc}
\toprule
\textbf{Model} & \textbf{100} & \textbf{200} & \textbf{300} & \textbf{500} & \textbf{1000} \\
\midrule
ResNet-18        & 0.233 & 0.288 & 0.323 & 0.374 & 0.451 \\
VIB              & 0.258 & 0.302 & 0.345 & 0.393 & 0.484 \\
infoR--LSF       & 0.262 & 0.320 & 0.357 & \textbf{0.429} & \textbf{0.542} \\
\midrule
Ours (Joint)     & \textbf{0.280} & 0.321 & 0.361 & 0.412 & 0.498 \\
Ours (Two-Stage) & 0.262 & \textbf{0.324} & \textbf{0.368} & 0.409 & 0.519 \\
\bottomrule
\end{tabular}
}
\vspace{-0.3cm}
\end{table}

\begin{figure*}[tb]
    \centering
    \begin{minipage}[b]{0.43\textwidth}
        \centering
        \includegraphics[width=\linewidth]{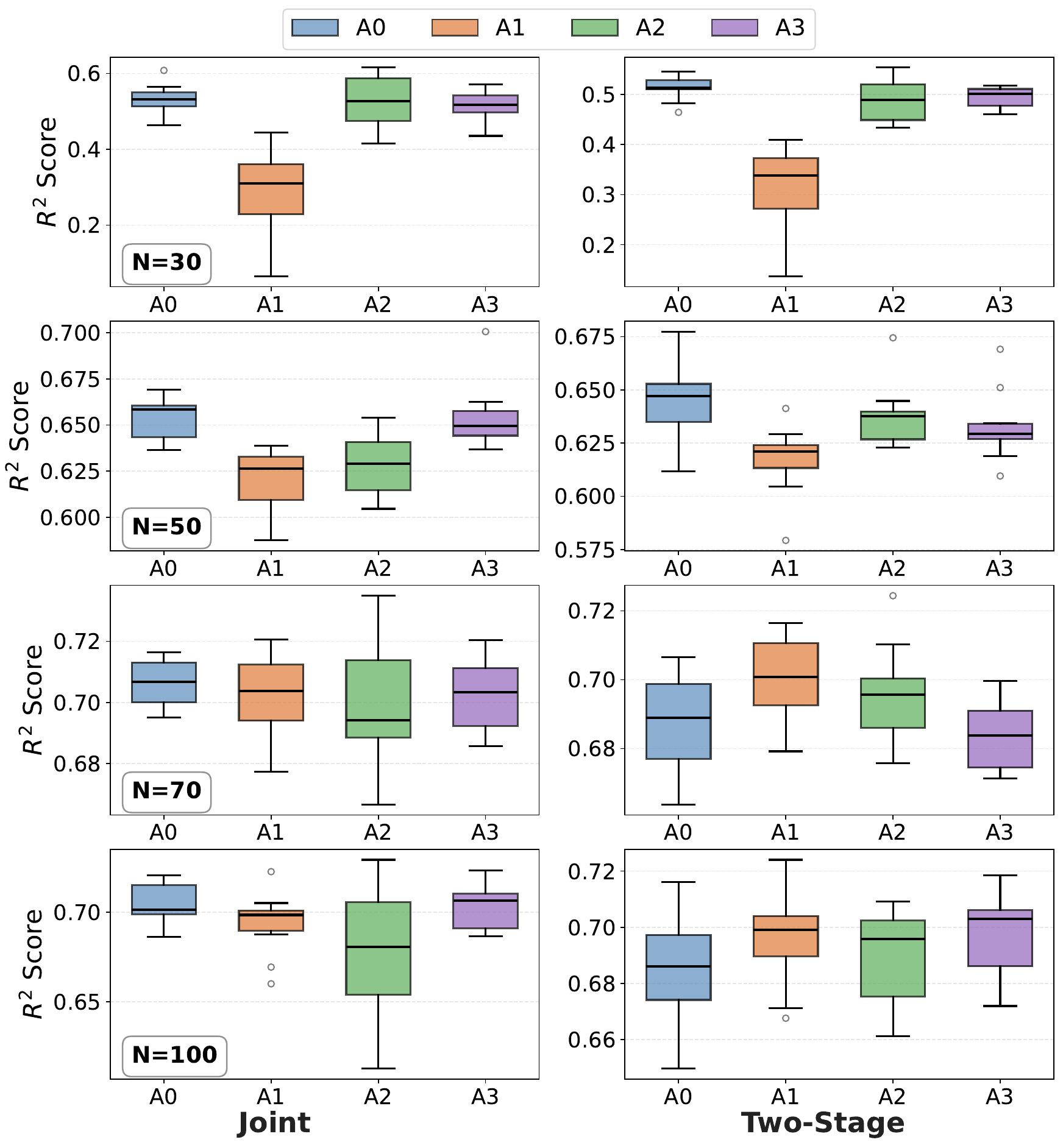}
        \centerline{\small \textbf{(a)} Concrete Dataset}
    \end{minipage}
    \hspace{0.5cm}
    \begin{minipage}[b]{0.43\textwidth}
        \centering
        \includegraphics[width=\linewidth]{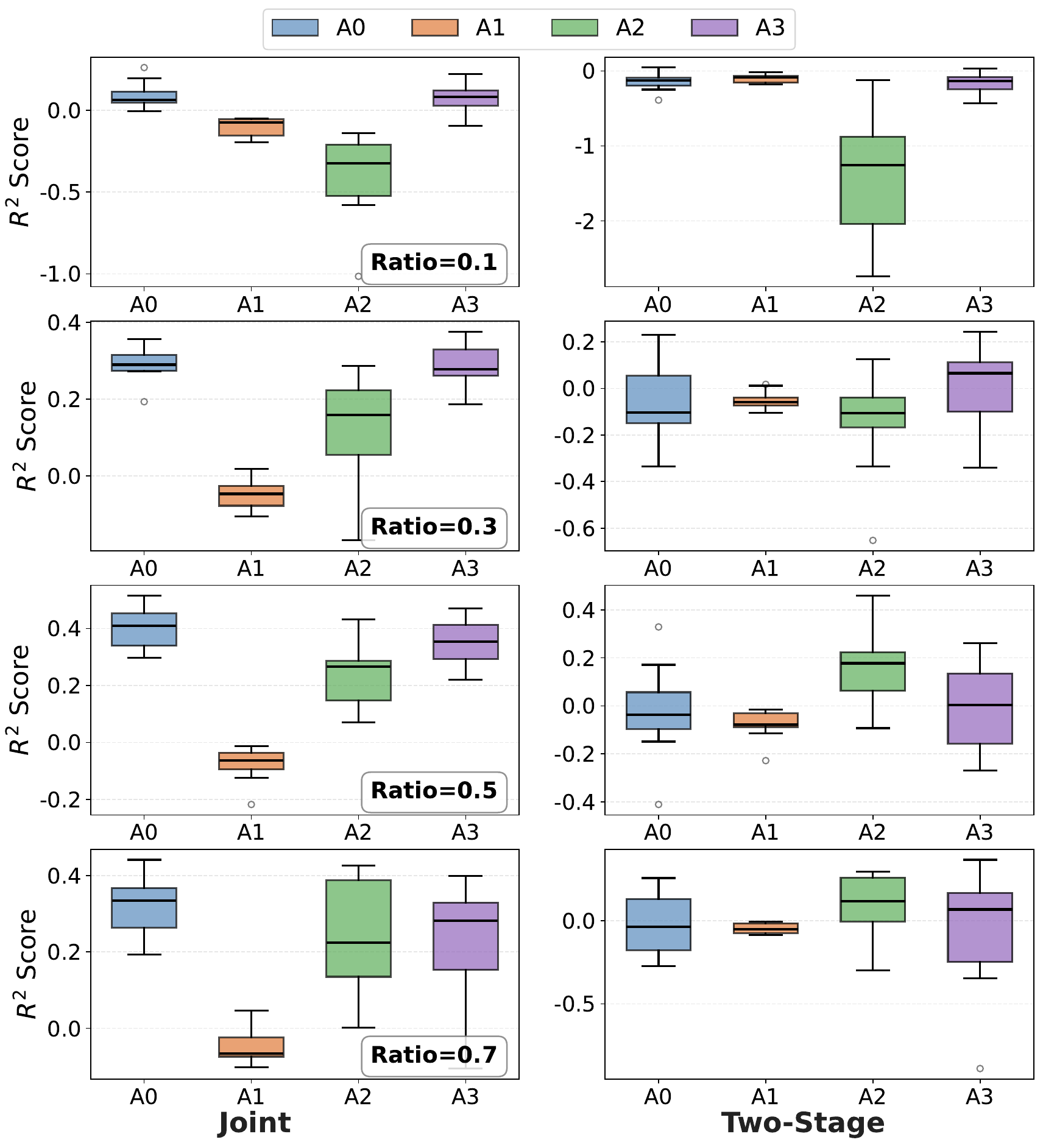}
        \centerline{\small \textbf{(b)} Synthetic Dataset}
    \end{minipage}
    \caption{\textbf{Ablation study results.} We compare the full AdverISF model (\textbf{A0} (Full Model)) against three variants: \textbf{A1} (w/o Multi-layer), \textbf{A2} (w/o Variational), and \textbf{A3} (w/o Adversarial). \textbf{(a)} Performance on the Concrete dataset across varying training set sizes ($N$). \textbf{(b)} Performance on the Synthetic dataset across varying training data ratios.}
    \label{fig:ablation_combined}
    \vspace{-0.3cm}
\end{figure*}

\subsection{Experimental Setup}
\label{sec:setup}
We evaluate AdverISF across diverse domains ranging from controlled synthetic environments to real-world applications.
For standard data-scarce regression benchmarks, we utilize the Concrete Compressive Strength dataset~\cite{concrete_compressive_strength_165} and the Appliance Energy Prediction (AEP) dataset~\cite{appliances_energy_prediction_374} from the UCI Machine Learning Repository.
We also evaluate representation learning using CIFAR-10~\cite{krizhevsky2009learning} in a few-shot setting.
To verify the effectiveness of separating task-relevant information from noise, we employ a synthetic dataset constructed with explicit dominant and subtle factors.
Crucially, we validate our method on a challenging real-world material design task for Composite Silicate Cement. This task focuses on predicting compressive strength from complex formulations. It rigorously tests model robustness against experimental noise in specialized scientific domains.

We compare against three baselines. We include standard MLP and ResNet-18 as conventional methods, VIB~\cite{alemi2017deep} as a classic compression-based method, and infoR-LSF~\cite{xie2024information} as the state-of-the-art retention approach.
For our framework, we explore two training strategies. Joint training optimizes the entire architecture simultaneously, whereas the two-stage approach trains each separation block sequentially. And, we apply stochastic weight perturbation to the KL divergence terms to enhance stability and diversity (Sections~\ref{sec:comparison} and ~\ref{sec:realworld}).
We report the coefficient of determination ($R^2$) for regression and accuracy for classification, averaged over $10$ seeds. Detailed architectures are reported in the Appendix.

\subsection{Comparison Analysis}
\label{sec:comparison}

We analyze generalization performance across diverse domains to validate the adversarial separation mechanism. On standard benchmarks (Tables~\ref{tab:r2_concrete} and~\ref{tab:r2_aep}), AdverISF consistently outperforms baselines, with particularly large margins in extreme data-scarce regimes (e.g., $N=30$ for Concrete). Unlike VIB, which suppresses subtle features through uniform compression, our method explicitly preserves critical signals by separating noise. We validate this property on the Synthetic dataset (Table~\ref{tab:r2_synthetic}), where AdverISF maintains robustness against noise distribution shifts while baselines degrade. This demonstrates that our method learns invariant task-relevant features. Finally, in the high-dimensional CIFAR-10 few-shot task (Table~\ref{tab:cifar10_acc}), AdverISF effectively prevents overfitting to background textures, a common pitfall for standard models like ResNet-18. This ensures superior generalization even with minimal supervision.

\subsection{Ablation Analysis}
\label{sec:ablation}

We investigate the contribution of key components comparing the full model (\textbf{A0}) against variants \textbf{A1} (w/o Multi-layer), \textbf{A2} (w/o Variational), and \textbf{A3} (w/o Adversarial).
As shown in Fig.~\ref{fig:ablation_combined}, all variants share the same hyperparameter for \textbf{A0}.
The multi-layer architecture is most critical. \textbf{A1} suffers severe performance drops, showing the multi-layer design is essential for recycling subtle features discarded in earlier layers.
Without variational regularization, \textbf{A2} exhibits the largest variance and extreme outliers, particularly on synthetic data.
Removing adversarial separation (\textbf{A3}) leads to significantly higher variance in low-data regimes, showing that explicit noise disentanglement stabilizes learning.
For training strategies, the two-stage approach demonstrates superior stability compared to joint optimization in extreme data-scarce settings (e.g., $N=30$ or $Ratio=0.1$). Decomposing the learning process ensures dominant features are captured before refining subtle features, preventing overfitting with limited supervision.

\begin{figure}[htb]
    \centering
    \includegraphics[width=\linewidth]{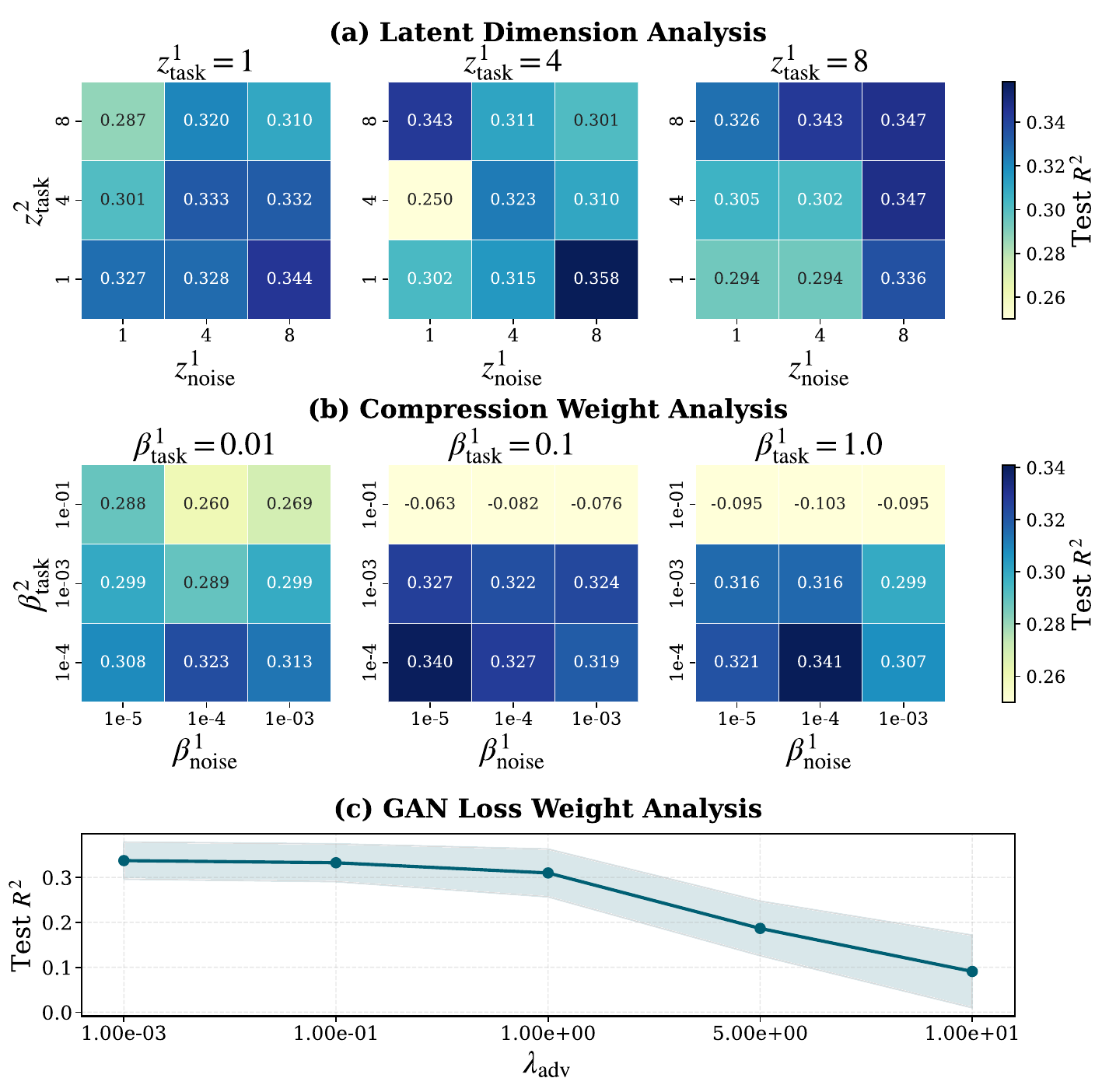}
    \caption{
        \textbf{Hyperparameter analysis on the Synthetic dataset.} 
        (a) Heatmaps showing $R^2$ stability across varying latent dimensions for layer 1 and layer 2.
        (b) Impact of KL divergence weights ($\beta$) on task and noise encoders.
        (c) Model performance robustness with respect to the adversarial loss weight $\lambda_{adv}$.
    }
    \label{fig:hyperparam}
\end{figure}

\begin{table}[htb]
\centering
\setlength{\tabcolsep}{11pt}

\caption{Performance comparison on the \textbf{Real Dataset} (70\% training data). Mean $R^2$ scores are reported. Best \textbf{Test} results are highlighted in \textbf{bold}.}
\label{tab:real_dataset}

\begin{tabular}{lccc}
\toprule
\textbf{Model} & \textbf{Train} & \textbf{Valid} & \textbf{Test} \\
\midrule
MLP              & 1.000 & 0.061 & -1.559 \\
VIB              & 0.999 & 0.812 & 0.617 \\
infoR--LSF       & 0.999 & 0.830 & 0.795 \\
\midrule
Ours (Joint)     & 0.976 & 0.808 & \textbf{0.897} \\
Ours (Two-Stage) & 0.976 & 0.724 & 0.882 \\
\bottomrule
\end{tabular}
\vspace{-0.3cm}
\end{table}

\subsection{Hyperparameter Analysis}
\label{sec:hyperparam}

We evaluate the stability of our hierarchical framework through hyperparameter sensitivity analysis on the Synthetic dataset. Fig.~\ref{fig:hyperparam}(a) shows that the model achieves optimal performance when the first task layer dimension is set to $z_{task}^1 = 4$. This configuration effectively balances information flow across hierarchical levels. An overly narrow bottleneck restricts representational power, while an excessively large dimension allows the primary layer to capture nearly all predictive information. Consequently, this prevents subsequent layers from functioning effectively and reduces the model to a standard VIB. Fig.~\ref{fig:hyperparam}(b) emphasizes the critical role of compression weights. Strong regularization on the secondary noise branch leads to performance degradation when primary task compression is high. This confirms that secondary branches require sufficient flexibility to retain subtle features that uniform compression might suppress. Finally, Fig.~\ref{fig:hyperparam}(c) demonstrates that generalization remains robust for adversarial weights $\lambda_{adv}$ up to 1.0. Beyond this threshold, $R^2$ scores decline sharply as extreme adversarial pressure forces the encoders to prioritize statistical independence over vital predictive information.

\subsection{Real-world Application: Composite Silicate Cement Design}
\label{sec:realworld}

We apply AdverISF to a challenging real-world task in Composite Silicate Cement (CSC) mix design. The goal is to predict compressive strength based on 8 formulation variables. These include curing age, water-cement ratio, and the contents of clinker, gypsum, slag, fly ash, steel slag, and limestone powder. To evaluate generalization in a realistic extrapolation scenario, we collected 60 historical formulations (randomly split 7:3 for training/validation) and conducted 9 new independent experiments for testing. This setup rigorously tests the model's ability to extrapolate to unseen recipes.

As shown in Table~\ref{tab:real_dataset}, the MLP baseline fails to generalize and yields a negative $R^2$ score of $-1.559$. This suggests that it captures spurious correlations rather than robust relationships between formulations and properties. The VIB model improves the performance to $0.617$ through input compression. Building on VIB, infoR-LSF reaches $0.795$ by maximizing feature retention. However, this method inadvertently preserves experimental noise alongside useful features, which limits its extrapolation performance. In contrast, AdverISF outperforms all baselines by a significant margin. Our Joint strategy achieves an $R^2$ of $0.897$ and the Two-Stage strategy reaches $0.882$. These results confirm that adversarial separation effectively decouples domain-relevant features from experimental noise. This decoupling enables much more robust property predictions for real-world material design.

\section{Conclusion}

In this work, we propose the paradigm of adversarial information separation, which isolates task-relevant information from confounding noise without explicit supervision.
We introduce AdverISF, a multi-layer framework that enforces statistical independence through a self-supervised adversarial mechanism. To further refine this separation, it progressively recycles noise representations across layers to recover subtle features that might otherwise be discarded.
We evaluate AdverISF on diverse benchmarks and a challenging real-world material design task, demonstrating superior performance in data-scarce and out-of-distribution scenarios. Ablation studies verify that our approach effectively distinguishes genuine generalizable features from structured noise, capturing subtle predictive patterns that are typically suppressed by uniform compression baselines.

\section*{Impact Statements}
This paper presents work whose goal is to advance the field of machine learning. There are many potential societal consequences of our work, none of which we feel must be specifically highlighted here.


\bibliography{example_paper}
\bibliographystyle{icml2026}

\newpage
\appendix
\onecolumn

\appendix

\section{Theoretical Analysis of Adversarial Information Separation}
\label{app:theory}

In this section, we analyze the theoretical guarantees of the proposed adversarial separation mechanism. We demonstrate that enforcing statistical independence between task features $z_{task}$ and noise representations $z_{noise}$ can be theoretically achieved through either Jensen-Shannon Divergence (Vanilla GAN) minimization or Wasserstein distance (WGAN) minimization.

\subsection{Independence via Jensen-Shannon Divergence (Vanilla GAN)}

Following the standard adversarial framework ~\cite{goodfellow2014generative}, the discriminator $D$ distinguishes between the joint distribution $P_J = p(z_{task}, z_{noise})$ and the product of marginals $P_M = p(z_{task})p(z_{noise})$. The minimax objective is formulated as:
\begin{equation}
    \min_E \max_D V(D, E) = \mathbb{E}_{\mathbf{x} \sim P_J}[\log D(\mathbf{x})] + \mathbb{E}_{\mathbf{y} \sim P_M}[\log(1 - D(\mathbf{y}))]
\end{equation}

\textbf{Proposition 1.} \textit{The global minimum of the standard GAN objective is achieved if and only if the Jensen-Shannon Divergence $D_{JSD}(P_J \| P_M) = 0$.}

Since $D_{JSD}(P \| Q) = 0$ implies $P = Q$ almost everywhere, optimizing this objective theoretically enforces $P_J = P_M$, satisfying the definition of statistical independence.

\subsection{Independence via Wasserstein Distance (WGAN)}

Alternatively, utilizing the Kantorovich-Rubinstein duality ~\cite{arjovsky2017wasserstein}, the separation can be formulated using a critic $D$ constrained to the set of 1-Lipschitz functions ($\|D\|_L \le 1$):
\begin{equation}
    \min_E \max_{D: \|D\|_L \le 1} \left( \mathbb{E}_{\mathbf{x} \sim P_J}[D(\mathbf{x})] - \mathbb{E}_{\mathbf{y} \sim P_M}[D(\mathbf{y})] \right)
\end{equation}

\textbf{Proposition 2.} \textit{Optimizing the WGAN objective is equivalent to minimizing the Wasserstein-1 distance $W_1(P_J, P_M)$.}

Since $W_1$ is a valid metric on the space of probability distributions, convergence $W_1 \to 0$ implies weak convergence $P_J \xrightarrow{d} P_M$, thereby guaranteeing $z_{task} \perp z_{noise}$.

While both frameworks theoretically ensure independence, we adopt WGAN-GP~\cite{gulrajani2017improved} for our implementation. The Wasserstein metric provides consistent, non-vanishing gradients even when high-dimensional latent distributions have disjoint supports, offering superior training stability in our data-scarce domains compared to the saturation-prone JS divergence.

\subsection{Hierarchical Feature Recovery via Conditional Mutual Information}
\label{app:multilayer_theory}

While the adversarial mechanism is highly effective at purifying task-relevant features, the strictness of the independence constraint introduces an inherent limitation regarding feature completeness. As noted in Section 3.4, the requirement to rigorously satisfy the independence condition $I(z_{task}; z_{noise}) \to 0$ compels the encoder to prioritize task-relevant features that are statistically distinct from noise. Consequently, ambiguous or subtle features bearing structural similarities to noise are inadvertently suppressed to avoid adversarial penalties.

This phenomenon results in the first separation layer functioning as a stringent filter that yields a pure but potentially incomplete task representation $z_{task}^{1}$. The noise representation $z_{noise}^{1}$ therefore absorbs the residual subtle information as an unavoidable byproduct of the separation process. To address this limitation, we provide an information-theoretic justification for the proposed Multi-layer Separation Architecture and demonstrate its capability to recover these hierarchical features through conditional mutual information maximization.

Let the total task-relevant information in input $x$ be decomposed into dominant features $F_{dom}$ and subtle features $F_{sub}$, such that $I(x; y) = I(F_{dom}; y) + I(F_{sub}; y \mid F_{dom})$.

\noindent\textbf{Step 1: Information Leakage in Layer 1.}
The first separation block minimizes $I(z_{task}^{1}; z_{noise}^{1})$ while maximizing the task prediction $I(z_{task}^{1}; y)$. Simultaneously, the Noise Extractor maximizes the conditional mutual information $I(z_{noise}^{1}; y \mid z_{task}^{1})$. 

Under the conservative separation strategy, the model prioritizes $F_{dom}$ for the task branch. Consequently, the subtle features are effectively ``crowded out'' and forced into the noise representation by the joint prediction objective:
\begin{equation}
    z_{task}^{1} \approx F_{dom}, \quad z_{noise}^{1} \supseteq F_{sub}
\end{equation}
This confirms that $z_{noise}^{1}$ is not pure noise but contains residual task information, \textit{i.e.}, $I(z_{noise}^{1}; y \mid z_{task}^{1}) > 0$.

\noindent\textbf{Step 2: Recovery via Conditional Maximization.}
The second layer takes $z_{noise}^{1}$ as input to extract a new representation $z_{task}^{2}$. Since $z_{task}^{1}$ is fixed, optimizing the second layer is equivalent to maximizing the mutual information from the residual space. By the Chain Rule of Mutual Information, the total information extracted is:
\begin{equation}
    I(z_{task}^{1}, z_{task}^{2}; y) = I(z_{task}^{1}; y) + \underbrace{I(z_{task}^{2}; y \mid z_{task}^{1})}_{\text{Information Gain}}
\end{equation}

\noindent\textbf{Proposition 3 (Strict Information Gain).} \textit{If the first layer is incomplete (i.e., information leakage $I(z_{noise}^{1}; y \mid z_{task}^{1}) > 0$) and the second layer has sufficient capacity, then the multi-layer architecture strictly increases the total task-relevant information:}
\begin{equation}
    I(z_{task}^{1}, z_{task}^{2}; y) > I(z_{task}^{1}; y)
\end{equation}

\textit{Proof Sketch.} Since $z_{task}^{2}$ is derived from $z_{noise}^{1}$ to maximize the prediction of $y$, it targets predictive signals orthogonal to $z_{task}^{1}$ (those classified as ``noise'' in Step 1). Thus, the conditional mutual information term is positive, representing the successful recovery of subtle features $F_{sub}$ that were discarded in the single-layer bottleneck.

\section{Hyperparameter Sensitivity Analysis}
\label{app:sensitivity}

In this section, we provide a detailed sensitivity analysis regarding the latent space configurations ($z^1_{task}$, $z^1_{noise}$, $z^2_{task}$), the KL divergence coefficients ($\beta^1_{task}$,$\beta^1_{noise}$, $\beta^2_{task}$), and the adversarial loss weight ($\lambda_{adv}$). The goal is to empirically justify the architectural choices and hyperparameter settings used in our main experiments.

\begin{figure}[ht]
    \centering
    \includegraphics[width=0.75\textwidth]{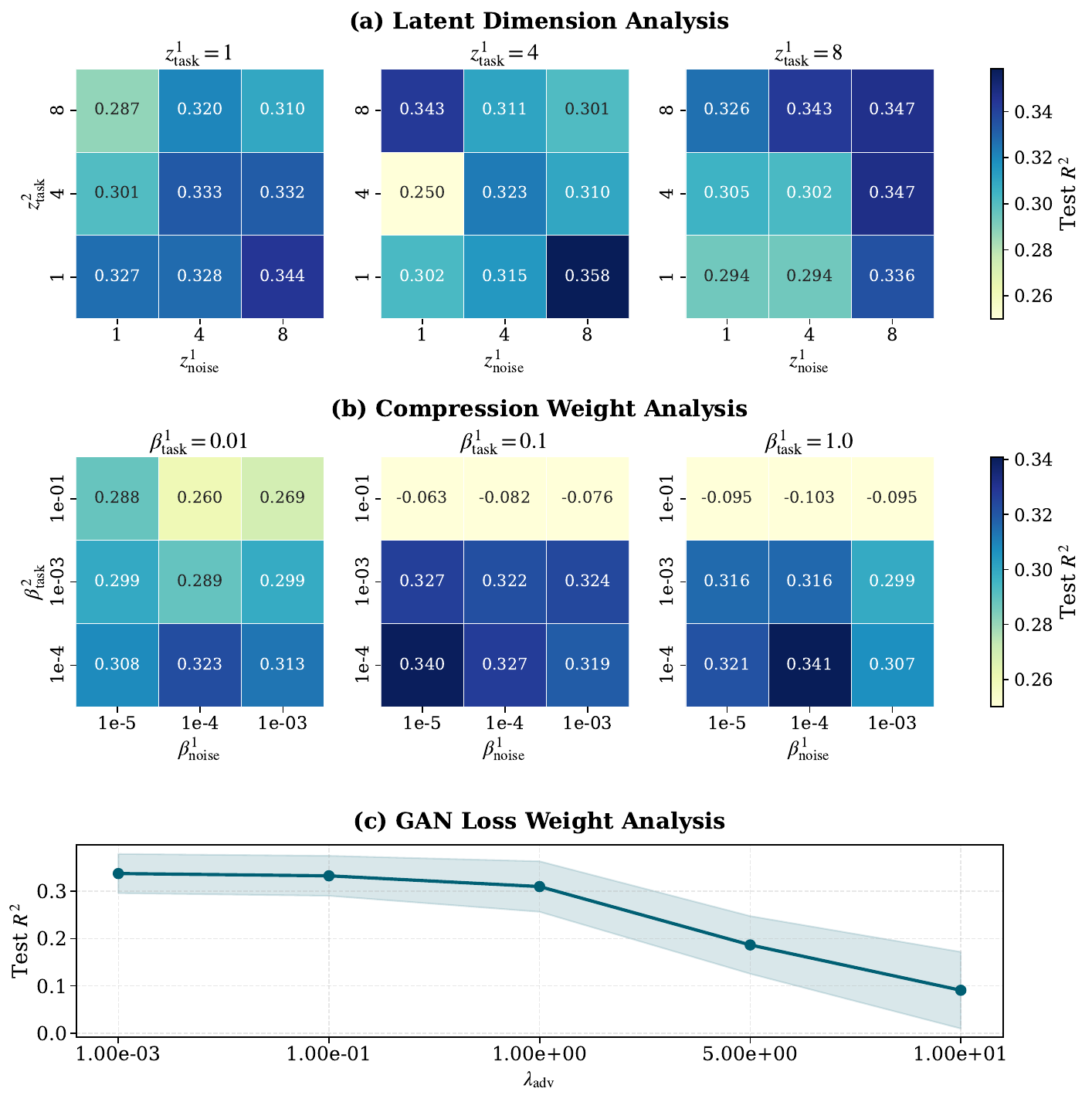}
    \caption{Hyperparameter sensitivity analysis on \textbf{Ratio 0.3} using \textbf{Joint Training}. The heatmap and curves illustrate performance stability across latent dimensions and weights.}
    \label{fig:ratio03_joint}
\end{figure}

\begin{figure}[htb]
    \centering
    \includegraphics[width=0.75\textwidth]{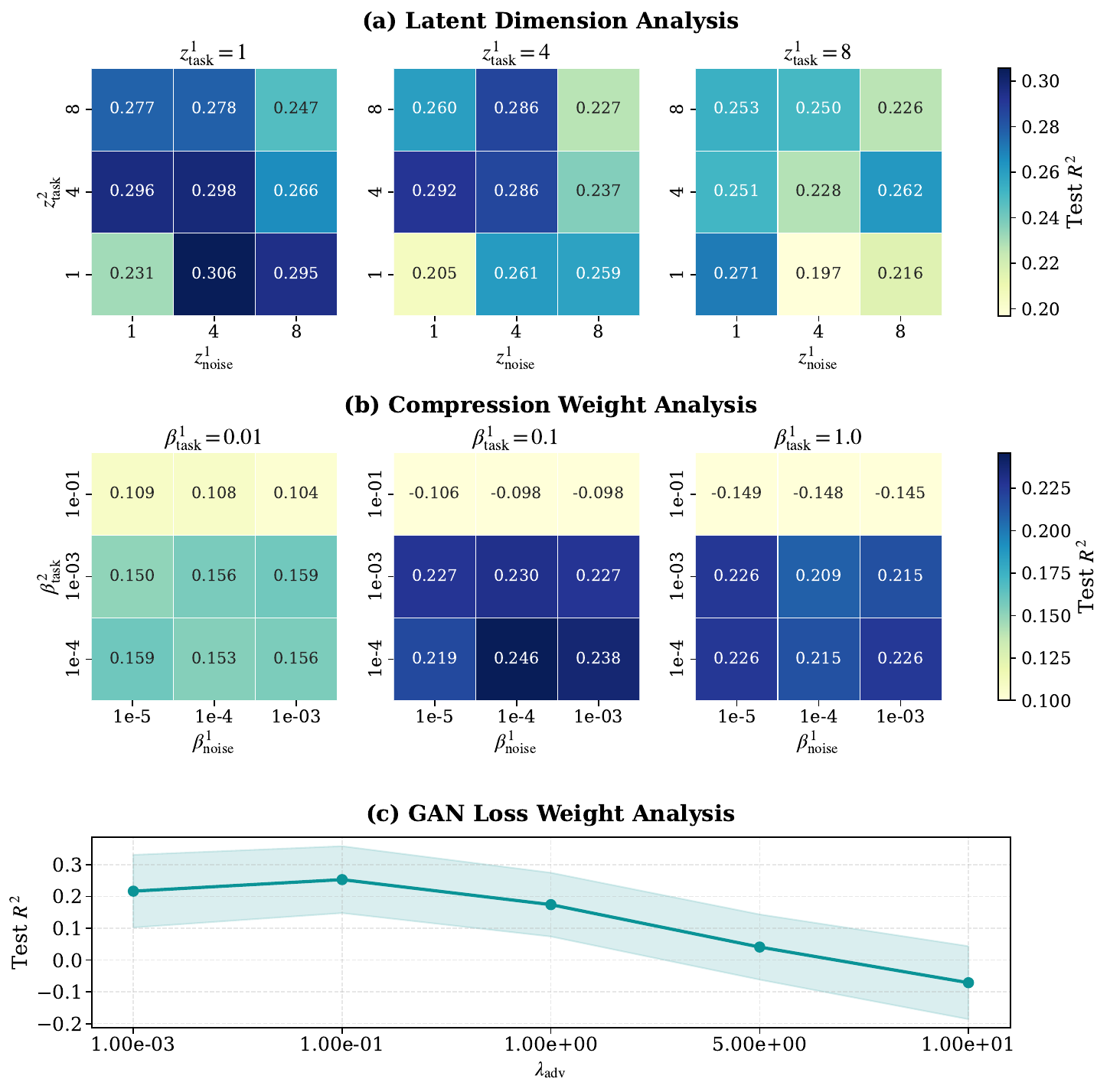}
    \caption{Hyperparameter sensitivity analysis on \textbf{Ratio 0.3} using \textbf{Two-Stage Training}. Comparison with Joint Training reveals different sensitivity patterns in the latent space.}
    \label{fig:ratio03_twostage}
\end{figure}

\begin{figure}[htb]
    \centering
    \includegraphics[width=0.75\textwidth]{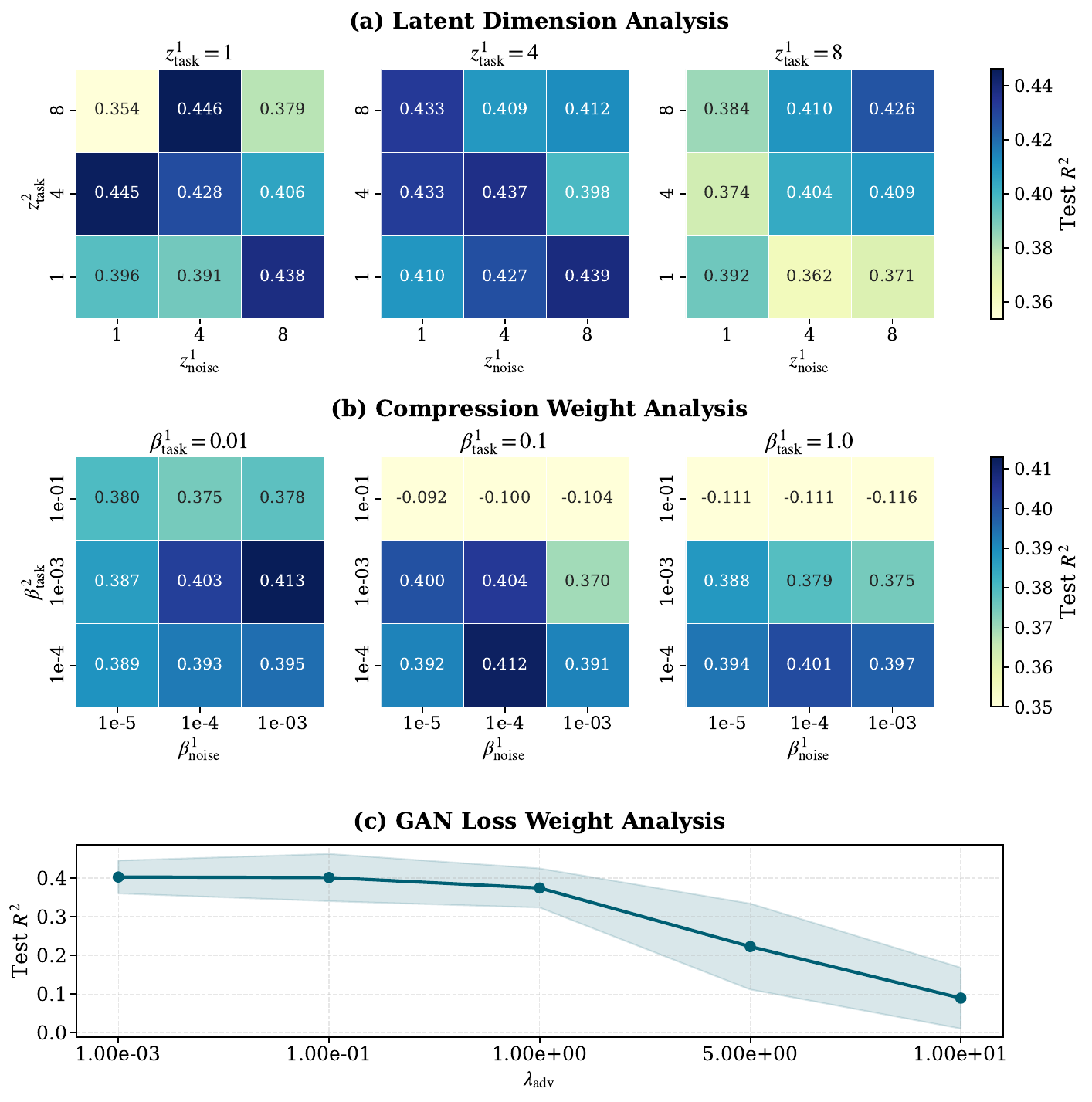}
    \caption{Hyperparameter sensitivity analysis on \textbf{Ratio 0.7} using \textbf{Joint Training}. With more training data, the model generally shows improved robustness in the sweet spot regions.}
    \label{fig:ratio07_joint}
\end{figure}

\begin{figure}[htb]
    \centering
    \includegraphics[width=0.75\textwidth]{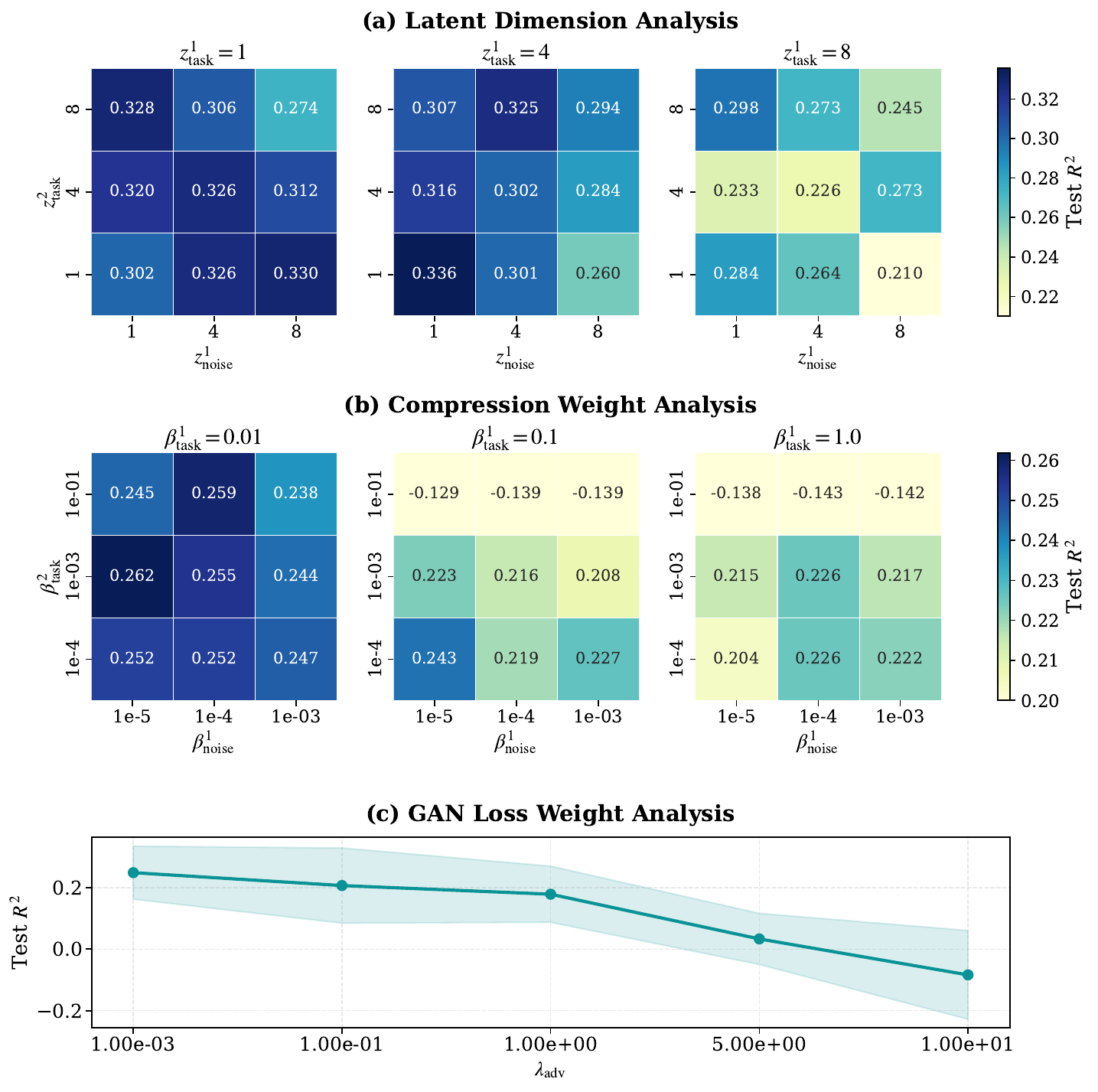}
    \caption{Hyperparameter sensitivity analysis on \textbf{Ratio 0.7} using \textbf{Two-Stage Training}. The wider confidence intervals at high $\lambda_{adv}$ indicate potential instability.}
    \label{fig:ratio07_twostage}
\end{figure}

\subsection{Impact of Latent Space Dimensionality}

To understand how the capacity of different latent components affects model performance, we conducted a grid search over the dimensions of the first-stage variables ($z^1_{task}$, $z^1_{noise}$) and the second-stage task variable ($z^2_{task}$), as visualized in the heatmaps of Figures~\ref{fig:ratio03_joint} through \ref{fig:ratio07_twostage}.

For the first-stage task variable $z^1_{task}$, we observed that compact dimensions (e.g., 1 or 4) yield superior performance. Surprisingly, increasing the dimension to 8 resulted in a general degradation of performance. We hypothesize that an overly large $z^1_{task}$ induces path dependence, where the model tends to encode all information—including noise and nuisances—into this single variable, rendering the subsequent hierarchical structure redundant. In this scenario, our model effectively degenerates into a standard Variational Information Bottleneck (VIB) baseline.

Conversely, for the noise variable $z^1_{noise}$, insufficient capacity leads to suboptimal results. A small dimension forces the model to discard complex nuisance variations or leak them into the task representation, thereby interfering with effective feature extraction. Therefore, the design of the first stage requires a trade-off: a compact task representation to force abstraction, paired with sufficient noise capacity to absorb task-irrelevant variations.

Regarding the second-stage task variable $z^2_{task}$, the model exhibits greater robustness. As the dimension varies from 1 to 8, performance fluctuations remain relatively mild, with intermediate values (e.g., 4) consistently achieving high stability (see Figures~\ref{fig:ratio03_joint} and \ref{fig:ratio07_joint}). Consequently, our final configuration adopts a compact $z^1_{task}$ to prevent model degeneration, while allocating adequate dimensions to $z^1_{noise}$ to ensure sufficient noise decoupling.

\subsection{Interplay of Compression Weights}

We further investigated the interaction effects of the compression weights(KL divergence weights): $\beta^1_{task}$, $\beta^1_{noise}$, and $\beta^2_{task}$. $\beta^1_{task}$ acts as the primary valve for the information bottleneck. Our results indicate that a smaller weight (e.g., $\beta^1_{task} \approx 0.01$) provides the most robust performance. As $\beta^1_{task}$ increases to 0.1 or 1.0, the model's tolerance for other hyperparameters decreases significantly. In high-compression regimes, the latent variables are forced to overly approximate the prior, leading to the loss of task-critical information.

A notable finding is the model's extreme sensitivity to the second-stage noise weight, $\beta^2_{task}$. Setting $\beta^2_{task}$ to a relatively large value (e.g., $10^{-1}$) causes a catastrophic drop in performance across all configurations, resulting in negative $R^2$ scores (visible as the light-colored regions in Figures~\ref{fig:ratio03_twostage} and \ref{fig:ratio07_twostage}). This suggests that excessive regularization at the second stage restricts the model's ability to model residuals, thereby blocking the reconstruction path. Conversely, a smaller weight ($10^{-3}$ or $10^{-4}$) provides the necessary flexibility for stable modeling.

The influence of $\beta^1_{noise}$ is relatively mild. Given reasonable settings for the task weights, $\beta^1_{noise}$ maintains good performance across the range of $10^{-5}$ to $10^{-3}$, with $10^{-4}$ often representing a local optimum. To prevent posterior collapse and ensure effective information flow, we recommend a light compression strategy: maintaining a minimal $\beta^2_{task}$ ($\le 10^{-3}$) combined with a moderately small $\beta^1_{task}$ ($0.01 - 0.1$).

\subsection{Adversarial Training Weight}

Finally, we evaluated the impact of the adversarial loss weight, $\lambda_{adv}$, on both generation quality and discriminative performance. We tracked the test set $R^2$ across different data ratios and training strategies.

The results reveal that adversarial training requires careful balancing. As shown in the line plots of Figure~\ref{fig:ratio03_joint} (Ratio 0.3) and Figure~\ref{fig:ratio07_joint} (Ratio 0.7), when $\lambda_{adv}$ is kept within a lower range (e.g., $10^{-3}$ to $10^{-1}$), the model demonstrates optimal stability and accuracy. However, as $\lambda_{adv}$ exceeds 1.0, performance deteriorates sharply across all configurations. At $\lambda_{adv} = 10.0$, the model effectively fails. This phenomenon indicates that excessive adversarial constraints disrupt the structural integrity of the latent space. The generator, in an attempt to deceive the discriminator, sacrifices the faithful reconstruction of the true data distribution, leading to mode collapse. Therefore, we recommend a mild adversarial regularization strategy ($\lambda_{adv} \le 0.1$) to enhance information separation without compromising critical task-relevant features.

\begin{table}[htb]
\centering
\caption{Detailed training configurations for baseline models. $d_z$ denotes the latent dimension for VIB. Values for InfoR-LSF on CIFAR-10 are aligned with its official implementation.}
\label{tab:baseline_settings}
\vspace{2mm}
\resizebox{0.75\textwidth}{!}{
\begin{tabular}{llcccccl}
\toprule
\textbf{Dataset} & \textbf{Method} & \textbf{Batch Size} & \textbf{Epochs} & \textbf{LR} & \textbf{Special Configs} \\
\midrule
    \multirow{3}{*}{AEP} 
    & MLP & 128 & 4000 & $3 \times 10^{-4}$ & - \\
    & VIB & 128 & 4000 & $3 \times 10^{-4}$ & $d_z=6, \beta=0.2$ \\
    & InfoR-LSF & 128 & 4000 & $3 \times 10^{-4}$ & - \\
    \midrule
    \multirow{3}{*}{Synthetic} 
    & MLP & 20 & 2000 & $3 \times 10^{-4}$ & - \\
    & VIB & 20 & 2000 & $3 \times 10^{-4}$ & $d_z=4, \beta=5 \times 10^{-4}$ \\
    & InfoR-LSF & 20 & 2000 & $3 \times 10^{-4}$ & - \\
    \midrule
    \multirow{3}{*}{Concrete} 
    & MLP & 20 & 4000 & $3 \times 10^{-4}$ & - \\
    & VIB & 20 & 4000 & $3 \times 10^{-4}$ & $d_z=4, \beta=0.05$ \\
    & InfoR-LSF & 20 & 4000 & $3 \times 10^{-4}$ & - \\
    \midrule
    \multirow{3}{*}{CIFAR-10} 
    & ResNet-18 & 256 & 200 & $5 \times 10^{-4}$ & - \\
    & VIB & 256 & 200 & $5 \times 10^{-4}$ & $d_z=64, \beta=0.1$ \\
    & InfoR-LSF & 256 & 200 & 0.1 & SGD$^\dagger$, WD=$5\times 10^{-4}$, StepLR \\
\bottomrule
\multicolumn{6}{l}{\footnotesize $^\dagger$ For CIFAR-10, InfoR-LSF uses SGD with momentum 0.9 and a step decay scheduler.}
\end{tabular}
}
\end{table}

\begin{table}[htb]
\centering
\caption{Hyperparameter configurations for our method. We report settings for both Joint and Two-Stage training. The latent space includes components $z_{task}^1$, $z_{task}^2$, and $z_{noise}^1$. KL weights for task representations follow a normal distribution $\beta \sim \mathcal{N}(\mu, \sigma^2)$. Fixed weights are reported as single values.}
\label{tab:ours_settings}
\vspace{2mm}
\setlength{\tabcolsep}{3.5pt}
\resizebox{\textwidth}{!}{
\begin{tabular}{llc cccccccc}
\toprule
\multirow{2}{*}{\textbf{Dataset}} & \multirow{2}{*}{\textbf{Method}} & \multirow{2}{*}{\textbf{Batch Size}} & \multirow{2}{*}{\textbf{Epochs}} & \multirow{2}{*}{\textbf{LR}} & \multirow{2}{*}{$\lambda_{adv}$} & \multicolumn{2}{c}{\textbf{Latent Dims}} & \multicolumn{3}{c}{\textbf{KL Weights} ($\mu / \sigma$)} \\
\cmidrule(lr){7-8} \cmidrule(lr){9-11} 
& & & & & & $z_{\text{task}}^1, z_{\text{task}}^2$ & $z_{\text{noise}}^1$ & $\beta^1_{\text{task}}$ & $\beta^2_{\text{task}}$ & $\beta^1_{\text{noise}}$ \\
\midrule
    \multirow{2}{*}{AEP} 
    & Joint &128 & 16000 & $2 \times 10^{-4}$ & 1.0 & 3, 3 & 5 & $0.60 / 0.05$ & $0.15 / 0.02$ & $8 \times 10^{-4}$ \\
    & Two-Stage &128 & $8000 \times 2$ & $3 \to 2 \times 10^{-4}$ & 1.0 & 3, 3 & 5 & $0.27 / 0.05$ & $0.15 / 0.02$ & $8 \times 10^{-4}$ \\
    \midrule
    
    \multirow{2}{*}{Synthetic} 
    & Joint &20 & 4000 & $2 \times 10^{-4}$ & 1.0 & 2, 2 & 5 & $0.25 / 0.001$ & $4 \times 10^{-4} / 1 \times 10^{-5}$ & $8 \times 10^{-5}$ \\
    & Two-Stage &20 & 3000 + 2000 & $3 \to 2 \times 10^{-4}$ & 1.3 & 2, 2 & 5 & $0.06 / 0.001$ & $4 \times 10^{-4} / 1 \times 10^{-5}$ & $8 \times 10^{-5}$ \\
    \midrule
    
    \multirow{2}{*}{Concrete} 
    & Joint &20 & 4000 & $2 \times 10^{-4}$ & 1.0 & 2, 2 & 5 & $0.25 / 0.01$ & $0.05 / 0.001$ & $8 \times 10^{-5}$ \\
    & Two-Stage &20 & $4000 \times 2$ & $3 \to 2 \times 10^{-4}$ & 1.3 & 2, 2 & 5 & $0.06 / 0.005$ & $0.005 / 5 \times 10^{-4}$ & $8 \times 10^{-5}$ \\
    \midrule
    
    \multirow{2}{*}{CIFAR-10} 
    & Joint &256 & 200 & $3 \times 10^{-4}$ & 1.0 & 32, 32 & 250 & $0.80 / 0.03$ & $0.15 / 0.01$ & $8 \times 10^{-4}$ \\
    & Two-Stage &256 & $200 \times 2$ & $3 \times 10^{-4}$ & 1.0 & 32, 32 & 250 & $0.60 / 0.03$ & $0.10 / 0.005$ & $8 \times 10^{-4}$ \\
\bottomrule
\end{tabular}
}
\end{table}

\section{Implementation Details}
\label{app:implementation}

In this section, we provide the network architectures and detailed hyperparameter configurations used across our experiments.

\subsection{Synthetic Dataset Construction Details}
\label{sec:synthetic_details}

To validate the model's ability to extract hierarchical information from high-dimensional data, we construct a synthetic dataset generated from a common latent source. We employ a non-linear projection mechanism to derive task-relevant latent factors.
The data generation pipeline consists of three stages: source sampling, latent feature projection, and target/observation generation.

\textbf{Source Sampling.}
We begin by sampling a high-dimensional source vector $s \in \mathbb{R}^{D_{total}}$ from a standard normal distribution $\mathcal{N}(0, I)$. This vector contains all the primitive stochastic information of the system. We set $D_{total}=13$.

\textbf{Latent Feature Projection.}
To simulate the complex extraction of meaningful features from raw data, we map the source vector $s$ to a semantic latent space using a randomly initialized Multi-Layer Perceptron (MLP), referred to as the \textit{Encoder}:
\begin{equation}
z_{base} = \text{MLP}_{enc}(s)
\end{equation}
where $z_{base} \in \mathbb{R}^{d_1 + d_2}$. We then partition this feature vector into two distinct components:
\begin{enumerate}[leftmargin=*]
\item \textbf{Dominant Factors ($z_{dominant}$)}: The first $d_1=3$ dimensions of $z_{base}$, representing high-level core features.
\item \textbf{Subtle Factors ($z_{subtle}$)}: The subsequent $d_2=5$ dimensions, representing fine-grained details.
\end{enumerate}
This projection ensures that the latent factors are non-linear transformations of the original source, creating a more challenging manifold learning task.

\textbf{Hierarchical Target Generation.}
The regression target $y$ is derived from these projected factors using two separate random MLPs, $f_{dom}$ and $f_{sub}$:
\begin{equation}
y = f_{dom}(z_{dominant}) + \gamma \cdot f_{sub}(z_{subtle}) + 0.15 \cdot \epsilon_{y}
\end{equation}
We set $\gamma=0.2$ to scale down the contribution of subtle factors, making them harder to detect.

\textbf{Observation Entanglement.}
The model observes a transformed version of the raw source vector $s$. The input observation $x$ is generated via a linear mixing transformation of the source:
\begin{equation}
x = \mathbf{W}_{trans} \cdot s + \mathbf{b}_{trans}
\end{equation}
where $\mathbf{W}_{trans} \in \mathbb{R}^{D_{total} \times D_{total}}$ is a random matrix. This implies that $x$ contains all the information present in $s$, but the semantic features $z_{dominant}$ and $z_{subtle}$ are implicit and must be recovered by inverting the relationship between $x$ and $y$.

\subsection{Network Architectures}

\textbf{Feature Backbones.}
For tabular datasets (AEP, Concrete, Synthetic, Cement), we employ Multi-Layer Perceptrons (MLPs) with LeakyReLU activations. For the CIFAR-10 image dataset, we adopt a ResNet-18 backbone with the final fully-connected layer removed to extract latent features.

\textbf{Model-Specific Configurations.}
\begin{itemize}
\item \textbf{Ours.} For tabular tasks, the encoders and predictors are MLPs with 2 hidden layers of sizes $[100, 100]$. The discriminator is an MLP with 2 hidden layers of sizes $[50, 50]$, taking the concatenation of task and noise representations as input. To ensure training stability, we implement the adversarial component using the Wasserstein GAN with Gradient Penalty (WGAN-GP) framework, where the discriminator (critic) is updated $n_{critic}=2$ times per generator iteration. For CIFAR-10, the encoder utilizes the ResNet-18 backbone, while the remaining components follow the MLP-based design.

To enhance training stability and representation diversity, we apply a stochastic weight perturbation strategy to the KL divergence terms. Instead of using fixed coefficients, we sample the penalty weight $\beta$ for each KL term from a normal distribution $\mathcal{N}(\mu, \sigma^2)$ for every training batch. This randomized scaling forces the model to adapt to varying regularization strengths. This process encourages the discovery of more robust and disentangled features. The specific $\mu$ and $\sigma$ values are determined based on the hyperparameter configurations for each dataset.

\item \textbf{MLP.} We employ an MLP with 3 hidden layers of sizes $[400, 400, 400]$ for these baselines. Since these models operate without a latent bottleneck, we adopt a shallower architecture. To ensure a fair comparison, we increase the hidden dimension to 400 to maintain a total parameter count consistent with the proposed method.

\item \textbf{ResNet-18.} We adopt the ResNet-18 architecture as implemented in the InfoR-LSF repository.

\item \textbf{VIB.} The encoder and predictors are MLPs with 3 hidden layers with hidden dimensions of $[200, 200, 200]$. The output dimension of encoder equals the sum of the dimensions of the two MLPs for $z^1_{task}$ and $z^2_{task}$. This component maps the input to a $d_z$-dimensional latent space. For CIFAR-10, the encoder utilizes the ResNet-18 backbone, while the remaining components follow the MLP-based design.

\item \textbf{InfoR-LSF.} We strictly follow the default architecture and hyperparameter settings from the official repository.
\end{itemize}

\subsection{Hyperparameter Settings}

\textbf{Baseline Configurations.}
Table~\ref{tab:baseline_settings} summarizes the training settings for each baseline. Unless otherwise noted, all models are optimized using the Adam optimizer. For InfoR-LSF on CIFAR-10, we adopt the original SGD configuration to maintain consistency with its reported performance.

\textbf{Ours Configurations.}
Table~\ref{tab:ours_settings} details the hyperparameters for our proposed method, including the specific distributions used for randomized KL weight sampling.

\subsection{Ablation Study Setup}
\label{sec:ablation_setup}

To verify the effectiveness of our proposed mechanisms, we design a comprehensive ablation study comparing four model variants and two training strategies.

\textbf{Model Variants.}
We establish the proposed architecture as the baseline and systematically disable key components to isolate their contributions:
\begin{itemize}
    \item \textbf{A0 (Full Model):} The complete \textbf{AdverISF} utilizing all three latent subspaces ($z^1_{task}, z^1_{noise}, z^2_{task}$) with both variational inference and adversarial disentanglement enabled.
    \item \textbf{A1 (w/o Multi-layer):} To evaluate the benefit of the multi-layer separation architecture, we disable the second layer. The model is restricted to a shallow capacity, freezing the second layer's parameters and generating predictions solely from the first level.
    \item \textbf{A2 (w/o Variational):} To test the impact of stochastic regularization, we remove the variational bottleneck. The reparameterization trick is disabled, and the KL divergence terms ($\beta$) are omitted.
    \item \textbf{A3 (w/o Adversarial):} To assess the role of noise disentanglement, we remove the discriminator and the adversarial loss ($\gamma_{adv}$). The model is trained purely on reconstruction and variational objectives.
\end{itemize}

\textbf{Optimization Strategies.}
We compare two training schedules to address stability in low-data settings:
\begin{enumerate}
    \item \textbf{Joint Training:} A standard end-to-end approach where all latent levels are optimized simultaneously.
    \item \textbf{Two-Stage Training:} A stepwise locking mechanism. In Phase 1 (initial 40\% of epochs), the model optimizes the first level ($z^1_{task}, z^1_{noise}$) while freezing the second level ($z^2_{task}$). In Phase 2, the first level is frozen to focus exclusive optimization on refining the deep latent features ($z^2_{task}$).
\end{enumerate}

\textbf{Hyperparameter Configurations.}
To assess the individual contribution of each component, we employ a unified set of hyperparameters for the ablation analysis. The latent dimensions are fixed as $dim(z^1_{task})=4$, $dim(z^1_{noise})=8$, and $dim(z^2_{task})=5$ across all variants. The specific regularization weights and discriminator iterations ($n_{critic}$) are configured as follows:
\begin{itemize}
    \item \textbf{Synthetic Dataset:} The KL weights are $\beta^1_{task}=0.2$, $\beta^1_{noise}=4.5 \times 10^{-6}$, and $\beta^2_{task}=5.5 \times 10^{-4}$. The adversarial training employs $\gamma_{adv}=0.11$ with $n_{critic}=6$.
    \item \textbf{Concrete Dataset:} The KL weights are $\beta^1_{task}=0.035$, $\beta^1_{noise}=1.9 \times 10^{-11}$, and $\beta^2_{task}=4.4 \times 10^{-5}$. The adversarial training employs $\gamma_{adv}=0.17$ with $n_{critic}=3$.
\end{itemize}

\begin{table}[htb]
    \centering
    \caption{Hyperparameter settings for the Real-world Application.}
    \label{tab:cement_settings}
    \renewcommand{\arraystretch}{1.2}
    \resizebox{0.7\linewidth}{!}{
    \begin{tabular}{l|l}
    \toprule
    \textbf{Category} & \textbf{Value / Setting} \\
    \midrule
    \multicolumn{2}{c}{\textit{Training \& Architecture}} \\
    \midrule
    Optimizer & Adam (Learning Rate: $5 \times 10^{-4}$, Batch Size: 20) \\
    Training Schedule & 15000 Epochs (Patience: 500) \\
    MLP Hidden Dim & 30 (Reduced from 400) \\
    Latent Dimensions & $z_{task}^1 = 2$, $z_{task}^2 = 2$, $z_{noise}^1 = 5$ \\
    \midrule
    \multicolumn{2}{c}{\textit{Method-Specific Coefficients}} \\
    \midrule
    \textbf{VIB} & $d_z=4$, $\beta = 0.2$ \\
    \midrule
    \multirow{3}{*}{\textbf{Ours (Joint)}} & $\lambda_{adv} = 1.0$ \\
    & $\beta^1_{task} \sim \mathcal{N}(\mu=1.5, \sigma=0.05)$, $\beta^2_{task} \sim \mathcal{N}(\mu=0.2, \sigma=0.01)$ \\
    & $\beta^1_{noise} = 8 \times 10^{-5}$ \\
    \midrule
    \multirow{3}{*}{\textbf{Ours (Two-Stage)}} & $\lambda_{adv} = 1.3$ \\
    & $\beta^1_{task} \sim \mathcal{N}(\mu=0.9, \sigma=0.1)$, $\beta^2_{task} \sim \mathcal{N}(\mu=0.1, \sigma=0.01)$ \\
    & $\beta^1_{noise} = 8 \times 10^{-5}$ \\
    \bottomrule
    \end{tabular}
    }
\end{table}

\subsection{Hyperparameter Analysis Setup}

We conduct a comprehensive ablation study on the synthetic dataset to evaluate the sensitivity of our model to structural and regularization hyperparameters. Experiments are performed under two data regimes: using $30\%$ and $70\%$ of the data for training.

\textbf{Training Strategies.}
We compare two optimization schedules:
\begin{enumerate}
    \item \textbf{Joint Training:} A standard end-to-end approach where all latent levels are optimized simultaneously for 5,000 epochs.
    \item \textbf{Two-Stage Training:} A stepwise approach designed to stabilize hierarchical learning. Phase 1 (0--2k epochs) optimizes $z^1_{task}$ and $z^1_{noise}$ while freezing $z^2_{task}$. Phase 2 (2k--5k epochs) freezes$z^1_{task}$ and $z^1_{noise}$ to focus exclusively on refining $z^2_{task}$.
\end{enumerate}

\textbf{Hyperparameter Configurations.}
We analyze performance variations across three axes. For robustness, each configuration is repeated over 10 independent trials.
\begin{itemize}
    \item \textbf{Latent Dimensions:} We vary the size of each latent level $z^1_{task}, z^1_{noise}, z^2_{task} \in \{1, 4, 8\}$ to test representation capacity.
    \item \textbf{Regularization Weights:} We assess the impact of the variational bottleneck and adversarial constraints by varying the KL divergence weights ($\beta^1_{task}, \beta^1_{noise}, \beta^2_{task}$ ranging from $10^{-5}$ to $1.0$) and the generator loss weight ($\gamma_{adv}$) ($GL$ from $0.001$ to $10.0$) with $n_{critic}=6$.
\end{itemize}

\subsection{Real-world Application Setup (Composite Portland Cement)}

\textbf{Dataset and Protocol.}
The dataset comprises 8 input variables: Curing Age, Water-Cement Ratio, and the mass percentages of six components. To strictly enforce a ``future prediction" setting, we utilized 60 historical samples for training and validation (randomly split 7:3) and reserved 9 samples synthesized in a separate, subsequent batch of experiments for testing. Inputs were normalized based on training set statistics to prevent leakage. Due to the extremely limited sample size, we reduced the hidden layer dimension of all MLPs from 400 to 30 to prevent overfitting.

\textbf{Implementation Details.}
We employed the Adam optimizer with early stopping. Detailed configurations for the training process, network dimensions, and method-specific hyperparameters are summarized in Table~\ref{tab:cement_settings}.


\end{document}